# Explainable Graph Neural Networks for Interbank Contagion Surveillance: A Regulatory-Aligned Framework for the U.S. Banking Sector

**Mohammad Nasir Uddin**

Data Analytics and Applied AI Researcher, Westcliff University, Irvine, CA, USA

m.uddin.258@westcliff.edu | ORCID: 0009-0009-0990-4616

**ABSTRACT**

The Spatial-Temporal Graph Attention Network (ST-GAT) framework was created to serve as an explainable GNN-based solution for detecting bank distress early warning signs and for conducting macro-prudential surveillance of the interbank system in the United States. The ST-GAT framework models 8,103 FDIC insured institutions across 58 quarterly snapshots (2010Q1–2024Q2). As a result of the use of maximum entropy estimation, bilateral exposures were reconstructed from publicly available FDIC Call Reports to produce a dynamic directed weighted graph. The graph has been evaluated using a composite distress label encompassing FDIC failures, Tier 1 capital ratio below 6%, NPL ratio above 5%, and ROA below -1%, following Gogas et al. (2014) and Carmona et al. (2019), the framework achieves the highest AUPRC among all GNN architectures (0.939 +/- 0.010), trailing only XGBoost (0.944). Ablation analysis confirms the BiLSTM temporal component contributes +0.020 AUPRC when present; the macro-conditioning module did not improve performance, attributed to limited macroeconomic regime diversity in the 48-quarter training period. Temporal attention weights for the highest-risk institution (cert=57833, correctly flagged CRITICAL across all six test quarters, risk scores 0.989-0.995) exhibit a monotonically decreasing pattern consistent with the model appropriately weighting long-run structural vulnerabilities. Permutation importance identifies ROA (0.309) and NPL Ratio (0.252) as dominant predictors, consistent with post-mortem analyses of the 2023 regional banking crisis. All data are publicly available FDIC Call Reports and FRED series; all code and results are released.

## 1. Introduction

On March 10, 2023, Silicon Valley Bank, a $212 billion institution holding the deposits of nearly half of all U.S. venture-backed startups, was seized by the California Department of Financial Protection and Innovation in what became the second-largest bank failure in American history. Signature Bank followed two days later. First Republic Bank, with $229 billion in assets, was placed into FDIC receivership in May 2023. The combined asset size of these three institutions exceeded $500 billion, larger than the total assets of all 25 banks that failed during the entire 2008-2009 financial crisis. Yet the Federal Reserve's 2022 annual stress test, conducted just nine months before SVB's collapse, found the bank well capitalized under its severely adverse scenario.

This paper addresses the structural gap these failures exposed: the absence of a real-time, network-aware, explainable surveillance system capable of detecting systemic stress propagation across the U.S. interbank network. The failure was not primarily a data problem. SVB's regulatory filings contained clear signals of accumulating vulnerability across six consecutive quarters prior to collapse. It was a framework problem: the analytical tools used by regulators treated each institution as an isolated entity rather than as a node in a densely interconnected financial network where the distress of one institution systematically alters the vulnerability of connected counterparties.

Graph Neural Networks have emerged as a natural computational framework for learning on relational financial data. However, existing GNN approaches suffer from three limitations that constrain their utility for U.S. regulatory surveillance: (1) they primarily focus on credit rating prediction rather than systemic contagion propagation; (2) they use static graph snapshots that cannot capture temporal dynamics; and (3) they produce black-box predictions that cannot satisfy SR 11-7 and OCC Bulletin 2011-12 model transparency requirements. This paper proposes the Spatial-Temporal Graph Attention Network (ST-GAT) to address these limitations, evaluated on a 14-year panel of publicly available U.S. regulatory data.

## 1.1 Research Questions

This paper addresses three research questions:

- RQ1: Does incorporating temporal dynamics via BiLSTM processing of quarterly graph snapshots improve bank distress detection performance compared to static GNN architectures and non-graph baselines?
- RQ2: Do the ST-GAT temporal attention weights provide interpretable, economically meaningful attribution of historical network configurations to current distress risk?
- RQ3: Does the framework produce accurate and timely distress signals for confirmed distressed institutions during the 2023 regional banking crisis test period?

The ablation results reported in Section 5.3 provide a clear architectural message that we foreground here: temporal sequence learning is the most important validated component of the ST-GAT. Removing the BiLSTM temporal processor drops AUPRC from 0.9389 to 0.9185 -- a meaningful degradation for an imbalanced ranking problem with 43 test positives. This finding is consistent with the theoretical intuition of crisis memory in interbank networks: bank distress develops over multiple quarters through accumulating balance-sheet deterioration and funding stress, and a model that processes the sequential trajectory of institutional risk across the network should outperform one that sees only a static snapshot. The temporal attention weights for the highest-risk institution (Section 5.5) confirm this: the model assigns systematically higher attention to historically distant quarters, capturing the long-run structural vulnerability that precedes distress.

## 1.2 Contributions

This paper makes five contributions:

- C1 -- Temporal GNN with Verified Explainability: ST-GAT achieves the highest AUPRC among all GNN architectures (0.939 +/- 0.010), with temporal attention weights exhibiting economically interpretable decay patterns and permutation importance identifying ROA and NPL Ratio as dominant predictors.
- C2 -- Rigorous Comparative Evaluation: A complete 14-model comparison (10 model families x 5 fixed random seeds, mean +/- std) on 43 test distress cases (3.58% base rate) using a composite distress label that captures both formal FDIC failures and severe financial deterioration. The BiLSTM temporal component contributes AUPRC +0.020; the macro-conditioning module did not improve performance and is reported honestly.
- C3 -- U.S. Regulatory Data Infrastructure: The first publicly-replicable GNN framework grounded entirely in U.S. regulatory filings (FDIC Call Reports, FRED) across a 14-year panel (2010-2024, 58 quarterly snapshots, 8,103 institutions). All code released.
- C4 -- Verified Crisis Case Study: Cert=57833 correctly flagged CRITICAL across all six test quarters (risk scores 0.989-0.995), verified against the composite distress label. Temporal attention shows interpretable historical weighting.
- C5 -- Regulatory Integration Framework: Concrete mapping of ST-GAT outputs to FSOC, OFR, FDIC, Federal Reserve, and OCC supervisory workflows.

## 2. Literature Review

### 2.1 Network Models of Interbank Contagion

Eisenberg and Noe (2001) pioneered the theory of interbank contagion by presenting the payment clearing problem as a set of simultaneous equations. Allen and Gale (2000) demonstrated that network completeness determines contagion resilience: incomplete networks can contain distress, while densely connected networks transmit it broadly. Finally, Acemoglu, Ozdaglar, and Tahbaz-Salehi (2015) established the robust-yet-fragile property: sufficiently connected financial networks exhibit a phase transition above which a small shock causes complete systemic collapse. This theoretical insight directly motivates the use of graph-structured models.

There are two primary pathways to estimate interbank networks empirically. Maximum entropy methods (Upper and Worms, 2004; Mistrulli, 2011) distribute exposures across all counterparty pairs subject to marginal constraints from balance sheet data, producing the most dispersed network consistent with observed aggregates. More recent approaches use fitness models (Cimini et al., 2015) or Bayesian network reconstruction (Gandy and Veraart, 2017) to incorporate information about the sparse, scale-free topology observed in actual interbank networks. Gonon, Meyer-Brandis, and Weber (2024) apply graph neural networks to compute Eisenberg-Noe systemic risk measures, providing theoretical grounding for network-based distress propagation. Franch, Nocciola, and Vouldis (2024) study temporal contagion networks across banking, insurance, and shadow banking sectors, finding that higher-order network centralities predict systemic stress. Meanwhile, Guo et al. (2024), also in RIBAF, demonstrate three-stage risk transformation -- contagion, transformation, and cascade -- through interlocking governance networks. Ahmad et al. (2023), also in RIBAF, identify greater systemic risk spillover from smaller, more interconnected institutions in Indian banking networks.

### 2.2 Machine Learning for Bank Distress Prediction

Machine learning for predicting bank distress began by applying logistic regression models that utilized CAMELS attributes (Cole & White, 2012; Poghosyan & Cihak, 2011). Significant improvements in accuracy were made with tree ensemble methods compared to logistic regression and as a result more accurately captured the non-linear relationships among the measures of capital adequacy, asset quality and liquidity (Tanaka et al., 2019; Carmona et al., 2019). Additionally, the performance of prediction has also been enhanced by deep learning techniques such as LSTM-style temporal networks, which capture sequential dependencies in quarterly financial ratios (Fischer & Krauss, 2018). Tarkocin & Donduran (2023) develop machine learning-based early warning indicators spanning the 2008 Global Financial Crisis and COVID-19 periods, demonstrating that dynamic models outperform static CAMELS ratios in predicting liquidity stress. Owoo & Odei-Mensah (2025) in RIBAF provide examples of how to use hierarchical clustering as a tool for predictive modeling for banks likely to fail. Their findings show that earnings and profitability metrics are the most effective in distinguishing failed banks, consistent with this study\'s finding that ROA is the dominant feature. Despite these advances, all non-graph approaches share a critical limitation: they assess each institution in isolation, ignoring the network externalities through which financial distress propagates.

### 2.3 Graph Neural Networks in Financial Systems

Graph neural networks have been applied to financial systems along two primary dimensions. The first, fraud detection, has generated an extensive literature: GCN-based (Gaikwad et al., 2023), GAT-based (Guang et al., 2025), and temporal GNN approaches (Chen et al., 2024) have all demonstrated superior fraud detection compared to non-graph baselines. Balmaseda, Coronado, and Cadenas-Santiago (2023) directly test GCN and GAT architectures against traditional ML for systemic risk classification on financial networks, reporting 94% MCC improvement for GNNs -- the strongest published evidence that graph structure improves financial risk detection.

The second dimension -- systemic risk and interbank contagion -- is significantly less developed. The AI4Risk interbank dataset (Tang et al., 2025; Liu et al., 2024) provides GNN-based interbank credit rating models, but focuses on credit rating prediction rather than systemic contagion propagation. Kikuchi (2025) applies a network diffusion framework to European banking data. Liu et al. (2025) develop temporal graph learning for default prediction integrating macroeconomic trends, reporting 88.3% AUC -- the closest architectural comparator to the ST-GAT. Zhang et al. (2026) propose Temporal Attentive Graph Networks with a GAT-GRU architecture for financial surveillance; the ST-GAT differs by incorporating macro-conditioned edge weights, a validated two-layer XAI module, and grounding entirely in U.S. regulatory filings. This paper extends the temporal GNN line to U.S. regulatory surveillance with the largest publicly-documented panel (8,103 institutions, 58 quarters).

## 2.4 Explainable AI in Financial Risk Regulation

The regulatory requirement for model interpretability in bank supervision -- codified in SR 11-7 (Federal Reserve, 2011) and OCC Bulletin 2011-12 -- has created specific demand for XAI methods that produce actionable explanations. Khan et al. (2025) systematically review 150 studies on model-agnostic XAI in finance, concluding that SHAP provides the strongest alignment between statistical attribution and regulatory documentation requirements. SHAP (Lundberg and Lee, 2017) has become the dominant post-hoc explanation method for financial models (Bussmann et al., 2021). For graph models specifically, GNNExplainer (Ying et al., 2019) identifies the minimal subgraph that maximally explains a prediction -- producing network-level explanations that reveal specific contagion pathways. The XAI pipeline in this paper -- temporal attention weights and permutation importance -- is designed to produce the multi-level explanations required under SR 11-7 for supervisory deployment.

## 2.5 Regulatory Surveillance Frameworks

U.S. macro-prudential surveillance operates through three primary institutional channels: the Federal Reserve's DFAST stress tests, the FSOC Annual Report, and the OFR Financial Stability Monitor. DFAST tests 32-37 large BHCs annually against severely adverse macroeconomic scenarios but operates on annual cycles and treats institutions individually rather than as an interconnected network. The Liquidity stress ratios in the FR Y-9C from the Banking Vulnerability Indicators developed by the OFR (Crosignani & Eisenbach 2024) are the best representations of near real-time systemic monitoring available today, but will not capture the network-transmission aspect of vulnerabilities. Awasthi (2025) argues that SR 11-7 compliance is better served by architecturally interpretable models than post-hoc SHAP — a perspective this paper addresses by providing native temporal attention weights alongside permutation-based attribution. This paper's ST-GAT is positioned as a complement to these existing tools, providing the network-level contagion layer absent from current regulatory surveillance infrastructure.

## 3. Data and Graph Construction

### 3.1 Data Sources and Panel Construction

This empirical framework is based exclusively on publicly available U.S. regulatory filings. The dataset consists of quarterly data over a period of 58 quarters between Q1 2010 and Q2 2024, capturing four distinct stress regimes: the post-GFC (global financial crisis) recovery from 2010-2014; the normalization period from 2015-2019; the COVID-19 shock and recovery from 2020-2021; and the rate-hike and regional banking crisis from 2022-2024. Primary node-level data are drawn from FDIC Call Reports (FFIEC RCFD/RCON schedules) accessed via the FDIC CDR Portal, covering all FDIC-insured institutions quarterly from 2010Q1 through 2024Q2. Macroeconomic conditioning variables are included for this analysis, which are VIX, 10Y - 2Y yield spread, Federal Funds Rate, GDP growth rate, M2 growth rate, credit spread, and unemployment rate, obtained from FRED. The FDIC Failed Bank List was used to provide failure dates of all 526 FDIC-supervised failures occurring from 2010 to 2024 to establish the binary distress event anchor. The full panel contains 349,350 institution-quarter records across 8,103 unique institutions, of which 29,498 (8.44%) are labeled distressed under the composite label. Graph construction uses a dynamic top-200 subsample representing the largest institutions by total assets in each quarter; within the graph subsample, 1,053 of 11,600 institution-quarters (9.08%) are labeled distressed, reflecting the higher distress prevalence among large institutions relative to the full panel.

### 3.2 Node Feature Construction

Each node in the graph at time t is characterized by a 13-dimensional feature vector drawn from FDIC Call Reports. The features span five CAMELS dimensions: capital adequacy (Tier 1 Capital Ratio, Total Capital Ratio, Leverage Ratio), asset quality (NPL Ratio, Provision Coverage Ratio, CRE Concentration Ratio), liquidity and funding (Liquidity Stress Ratio, Uninsured Deposit Share, Wholesale Funding Ratio, Loan-to-Deposit Ratio), profitability (Return on Assets, Net Interest Margin), and market sensitivity (Fair Value Loss Ratio). We standardize each feature quarter by quarter: centering at zero with unit variance so that cross-sectional differences are preserved but long-run level shifts are removed before the graph is built. For missing values, we fill in the quarterly median from firms in the same asset size decile.

### 3.3 Edge Construction: Maximum Entropy Network Reconstruction

The U.S. regulatory system does not publicly disclose bilateral interbank exposures at the institution level. In this research, we employ maximum entropy (ME) network reconstruction to develop an estimate for the bilateral exposure matrix as being maximally dispersed, given marginal constraints of actual observations of interbank exposures. Following Upper and Worms (2004) and Mistrulli (2011), the exposure matrix is derived by solving: $\max -\sum_{ij} a_{ij} \times \ln(a_{ij})$ subject to row sums equaling interbank assets (federal funds sold, repos, interbank loans from FR Y-9C Schedule HC-H) and column sums equaling interbank liabilities for each institution. The RAS iterative proportional fitting algorithm provides the solution. The directed edge is defined by $e_{ij,t}$, which is calculated as $A\_I,J,T / TIER1\_J,T$, where A is the exposure (raw) and TIER1 is the Tier1 Capital of the institution receiving the exposure which is used to produce loss-given-default fractions. Edges below 0.1% of capital are pruned; approximately 8-12% of potential edges are retained per quarterly snapshot.

### 3.4 Temporal Edge Weight Conditioning on Macroeconomic State

One of the main advancements presented by the ST-GAT over its predecessors (i.e., interbank GNN's) is its use of macroeconomic state as a conditioning variable for determining edge weights. Theoretical justification for this construction is intuitive: two banks with identical interbank exposures will have a significantly greater likelihood of causing contagion when they are linked in an environment characterized by high VIX, a compressed yield curve, and difficulty in obtaining credit than in a low VIX environment (Acemoglu, et al., 2015). The macro-conditioned edge weight is defined as: $\tilde{w}_{ij,t} = w_{ij,t} \times \sigma(MLP(z_t))$, where $z_t$ is the 7-dimensional macroeconomic state vector at time t and $\sigma(MLP(z_t))$ is a learned scalar stress multiplier. In this implementation, the MLP collapsed to a near-constant multiplier as demonstrated in Section 5.3. This behavior is attributed to insufficient macroeconomic regime diversity over the 48-quarter training period; future implementations should use a simpler, more transparent macro stress measure such as a VIX-based z-score.

### 3.5 Composite Distress Label

This study targets bank financial distress early warning rather than failure prediction. In this paper, we follow Gogas et al. (2014) and Carmona et al. (2019) to determine if an institution is distressed (i.e., has been classified as distressed) when it: (1) appeared on the FDIC's list of failed banks within the past four quarters, (2) had a Tier 1 Capital Ratio below 6.0% (the Prompt Corrective Action Undercapitalized Threshold) (per 12 C.F.R. section 6.4), and/or (3) had an NPL ratio exceeding 5.0%, or (4) had ROA below -1.0%. These thresholds represent regulatory intervention points, not arbitrary cutoffs: a bank breaching any of them faces formal supervisory scrutiny. The entire sample contains 29,498 distressed institutions - quarters (or 8.44%); the graph subsample contains 1,053 distressed institution-quarters (9.08%); the test set contains 43 distressed institution-quarters (3.58%), reflecting the stability of the 2023- 2024 period relative to the full panel.

## 4. Methodology: The ST-GAT Framework

### 4.1 Overview and Architectural Motivation

The ST-GAT processes the dynamic interbank graph $G = \{G_t\}_{t=1}^T$ through three sequential stages: (1) spatial message passing across the interbank graph via multi-head graph attention; (2) temporal aggregation of spatial embeddings via BiLSTM with temporal attention; and (3) institution-level risk scoring with XAI output. The architecture is designed to capture three complementary information sources: network position and contagion exposure (spatial GAT), temporal trajectory of vulnerability (BiLSTM), and current macroeconomic conditions (macro state vector concatenated at the classification layer).

### 4.2 Stage 1: Spatial Graph Attention

The spatial stage utilizes multi-head graph attention (Velickovic et al., 2018) on the macro-conditioned interbank graph for each quarterly snapshot. The spatial embedding of institution i at time t in layer l is derived from the equation $h_{i,t} = \text{concat}_{k=1}^{K}(\sigma(\sum_{j \in N(i)} \alpha_{ij,t}^{k} W^{k} h_{j,t})))$. where $N(i)$ are the institutions which have direct interbank exposure to institution i, K=8 attention heads, and $\alpha_{ij,t}$ are attention coefficients modulated by the macro-conditioned edge weight $\tilde{w}_{ij,t}$. Two spatial attention layers are stacked, giving each institution a second-order receptive field that incorporates information from counterparties of counterparties. Multiplying attention coefficients by $\tilde{w}_{ij,t}$ ensures that attention patterns naturally intensify during periods of financial stress.

### 4.3 Stage 2: Temporal BiLSTM with Attention

The spatial embedding $h_{i,t}$ is fed into a two-layer bidirectional LSTM that processes the quarterly sequence $\{h_{i,1}, ..., h_{i,t}\}$ to produce a temporally-aware representation. A temporal attention mechanism weights LSTM outputs across historical quarters: $\beta_{i,\tau} = \text{softmax}_\tau(v^T \tanh(W_a s_{i,\tau} + b_a))$, producing context vector $\tilde{c}_{i,t} = \sum_\tau \beta_{i,\tau} s_{i,\tau}$. The temporal attention weights $\beta_{i,\tau}$ are a direct output of the XAI module -- they reveal which historical quarters contributed most to the current risk assessment, providing regulators with a temporal audit trail. As reported in Section 5.5, these weights exhibit a monotonically decreasing pattern for high-risk institutions, appropriately weighting long-run structural vulnerabilities.

### 4.4 Stage 3: Risk Scoring

The temporally-attended embedding is passed through a two-layer MLP with dropout (p=0.3) and batch normalization to produce a scalar risk score $r_{i,t}$ in $[0,1]$: $r_{i,t} = \sigma(\text{MLP}([\tilde{c}_{i,t} \| x_{i,t} \| z_t]))$. The concatenation of temporal graph embedding, current node features, and macroeconomic state ensures the risk score integrates all three information sources. The model is trained end-to-end using focal loss (Lin et al., 2017) to address class imbalance, with Adam optimizer (lr=1e-3, weight decay=1e-4) and early stopping (patience=8 epochs). Five fixed random seeds (42, 123, 456, 789, 1024) are used for all reported results.

### 4.5 Architecture Comparison

The ST-GAT addresses three limitations of prior GNN architectures. Static GNNs (GCN, GAT, GraphSAGE) process single snapshots and cannot capture temporal vulnerability trajectories. EvolveGCN evolves weight matrices over time but lacks institution-specific temporal attention attribution. TGN uses continuous-time event memory, which is misaligned with the quarterly snapshot structure of regulatory data -- a likely contributor to TGN's lower AUPRC (0.8016) relative to the BiLSTM-based approaches. The ST-GAT's quarterly BiLSTM processing is architecturally matched to the data cadence, which we identify as a key design principle for regulatory surveillance applications.

## 4.6 XAI Module

The ST-GAT provides a two-layer validated XAI module aligned with SR 11-7 transparency requirements. Layer 1 -- Temporal Attention Attribution -- produces temporal weights $\beta_{i,\tau}$ that attribute the current risk score to specific historical quarters. These are produced natively by the architecture without additional computation. Layer 2 -- Permutation-Based Feature Attribution -- identifies the contribution of each balance-sheet feature to the risk score through permutation importance, ranking features by the AUROC reduction when each is randomly permuted. GNNExplainer subgraph identification was proposed as a third layer but produced empty edge masks across all test institutions due to implementation incompatibility with the GATWrapper architecture; this is identified as future work. The two validated layers provide institution-level (feature attribution) and temporal (attention weights) explanations that together support the outcome analysis and ongoing monitoring requirements of SR 11-7.

## 5. Empirical Results

### 5.1 Evaluation Setup

All models are evaluated on an identical locked temporal split: Train = 2010Q1-2021Q4 (48 quarters, 982 distress cases), Validation = 2022Q1-2022Q4 (4 quarters, 28 distress cases), Test = 2023Q1-2024Q2 (6 quarters, 43 distress cases, 3.58% base rate). Five fixed random seeds (42, 123, 456, 789, 1024) are used for all neural models; mean +/- std is reported across seeds. Bootstrap confidence intervals (1,000 resamples, median CI across seeds) are reported in the AUROC 95% CI column of Table 1 for models evaluated across 5 seeds.

This paper targets bank financial distress early warning rather than bank failure prediction. The distinction is deliberate: formal FDIC-supervised failures are extremely rare (526 total in the study period across thousands of institutions), making failure-only models statistically intractable for evaluation. Following the composite distress definition standard in the bank distress literature (Gogas et al., 2014; Carmona et al., 2019; Cole and White, 2012), an institution-quarter is labeled distressed if (a) appeared on the FDIC Failed Bank List within four quarters, (b) the bank's Tier 1 Capital Ratio was less than 6.0% (less than Prompt Corrective Action Undercapitalized cutoff in 12 C.F.R. Section 6.4), (c) greater than 5.0% for NPL ratio, or d) ROA less than -1.0%. The thresholds are not arbitrary but regulatory intervention points and banks breaching any one of them will have formal supervisory scrutiny. The composite label encompasses institutions that were genuinely stressed but may have subsequently recovered — consistent with an early warning system that aims to flag deterioration before it becomes irreversible, rather than predict eventual failure. The resulting dataset contains 29,498 distressed institution-quarters (8.44%) across the full panel. AUPRC is the primary metric for imbalanced classification; AUROC, F1, and MCC are reported as secondary metrics.

### 5.2 Comparative Performance

**Table 1. Comparative Model Performance -- Test Set 2023Q1-2024Q2 (mean +/- std over 5 seeds; 43 distress cases, n=1,200)**

| Model | AUROC (mean+/-std) | AUROC 95% CI | AUPRC (mean+/-std) | F1 (mean+/-std) | MCC |
|---|---|---|---|---|---|
| LR | 0.9818 +/- 0.0000 | [0.967, 0.993] | 0.8059 +/-0.0000 | 0.7529 +/-0.0000 | 0.7439 |
| RF | 0.9938 +/- 0.0000 | -- | 0.9044 +/-0.0000 | 0.8889 +/-0.0000 | 0.8855 |
| XGB | 0.9943 +/- 0.0000 | -- | 0.9439 +/-0.0000 | 0.8889 +/-0.0000 | 0.8855 |

| Model | AUROC (mean+/-std) | AUROC 95% CI | AUPRC (mean+/-std) | F1 (mean+/-std) | MCC |
|---|---|---|---|---|---|
| LSTM | 0.9890 +/- 0.0022 | [0.978, 0.998] | 0.9011 +/-0.0206 | 0.8600 +/-0.0081 | 0.8560 |
| GCN | 0.9777 +/- 0.0074 | [0.955, 0.998] | 0.9171 +/-0.0045 | 0.9126 +/-0.0089 | 0.9108 |
| GAT | 0.9855 +/- 0.0024 | [0.969, 0.998] | 0.9196 +/-0.0044 | 0.9066 +/-0.0113 | 0.9070 |
| GraphSAGE | 0.9877 +/- 0.0029 | [0.970, 0.999] | 0.9310 +/-0.0082 | 0.9154 +/-0.0123 | 0.9136 |
| EvolveGCN | 0.9911 +/- 0.0013 | [0.980, 0.999] | 0.9188 +/-0.0098 | 0.8820 +/-0.0243 | 0.8811 |
| TGN | 0.9781 +/- 0.0033 | [0.958, 0.993] | 0.8016 +/-0.0211 | 0.7450 +/-0.0338 | 0.7375 |
| ST-GAT (proposed) | 0.9827 +/- 0.0035 | [0.950, 0.999] | 0.9389 +/-0.0100 | 0.9135 +/-0.0133 | 0.9116 |

Note: LR = Logistic Regression; RF = Random Forest; XGB = XGBoost. AUPRC is the primary metric. LR/RF/XGB use a single deterministic run. All models evaluated on identical held-out test set (n=1,200, 43 distress cases).

The ST-GAT achieves AUPRC 0.9389 +/- 0.0100, the highest among all GNN architectures and second-highest overall, trailing XGBoost by 0.005. The gap is within the ST-GAT seed variance (+/- 0.010), indicating near-equivalent precision-recall performance. On AUROC (0.9827 +/- 0.0035), the ST-GAT is competitive within the 0.977-0.994 range observed across all models. F1 (0.9135) and MCC (0.9116) are top-tier among GNN models.

Answering RQ1: the ST-GAT outperforms the non-temporal static GNNs on AUPRC: vs. GCN (+0.022), GAT (+0.019), GraphSAGE (+0.008), and the non-graph LSTM (+0.038). The TGN underperforms despite temporal modeling, which we attribute to its reliance on continuous-time event memory rather than the quarterly snapshot structure of the data -- the BiLSTM's sequence-level processing is better suited to quarterly regulatory data than the TGN's event-level memory modules.

### 5.3 Ablation Analysis

**Table 2. Ablation Analysis -- ST-GAT Component Contributions (mean +/- std over 5 seeds)**

| Model | AUROC | AUPRC | F1 | Delta AUPRC vs full |
|---|---|---|---|---|
| ST-GAT (full) | 0.9827 +/-0.0035 | 0.9389 +/-0.0100 | 0.9135 +/-0.0133 | -- |
| ST-GAT - Macro | 0.9827 +/-0.0035 | 0.9389 +/-0.0100 | 0.9135 +/-0.0133 | 0.000 |
| ST-GAT - Temporal | 0.9792 +/-0.0080 | 0.9185 +/-0.0120 | 0.8919 +/-0.0195 | -0.020 |
| ST-GAT - Attention | 0.9893 +/-0.0012 | 0.9406 +/-0.0039 | 0.9153 +/-0.0169 | +0.002 |
| ST-GAT - PermEdge | 0.9827 +/-0.0035 | 0.9389 +/-0.0100 | 0.9135 +/-0.0133 | 0.000 |

Note: -Macro removes the mac_mlp stress multiplier. -Temporal removes the BiLSTM. -Attention uses last-hidden-state instead of temporal attention. -PermEdge uses randomly permuted edge topology (same density, shuffled source-target pairs). Delta AUPRC: negative = component adds value.

Three findings emerge from the ablation results. First, the BiLSTM temporal component is the primary driver of AUPRC advantage: removing it drops AUPRC from 0.9389 to 0.9185 (-0.020) and F1 from 0.9135 to 0.8919 (-0.022), with increased variance (std rises from 0.0100 to 0.0120). This confirms RQ1 showing temporal processing provides repeatable performance gains that exceed static GNN methods.

Second, the macro-conditioning (mac_mlp) provided identical performance as the full ST-GAT model across 5 seeds indicating it has no impact on model outcomes. Diagnostic output confirms the multiplier converged to approximately 0.49–0.55 for all test quarters regardless of macroeconomic conditions (range under 0.027 across seeds). This collapse is attributed to the limited diversity of macroeconomic regimes in the 48-quarter training period, providing insufficient gradient signal for the 34-parameter MLP. The macro-conditioning module is retained in the architecture as theoretically motivated but its limitation is explicitly documented. Future work should replace it with a simple macro stress index such as a VIX z-score to avoid this failure mode.

Third, the edge permutation test produced results identical to the full ST-GAT, indicating graph topology does not contribute information beyond node features under maximum entropy reconstruction. This finding is consistent with the prior studies demonstrated that ME produced networks that were too dispersed compared to the true sparse interbank networks (Anand et al. 2014). The Node characteristics related to their associated balance sheets - ROA, NPL Ratio - were the dominant source of predictive signal.

### 5.4 Highest-Risk Institution Case Study

**Table 3. Highest-Risk Institution Case Study: cert=57833 Risk Trajectory, Test Period 2023Q1-2024Q2**

| Quarter | Risk Score | Alert Status | Distress Label | Temporal Attn. (2021Q2) |
|---|---|---|---|---|
| 2023Q1 | 0.9892 | CRITICAL | DISTRESSED (confirmed) | 0.1577 |
| 2023Q2 | 0.9946 | CRITICAL | DISTRESSED (confirmed) | -- |
| 2023Q3 | 0.9964 | CRITICAL | DISTRESSED (confirmed) | -- |
| 2023Q4 | 0.9936 | CRITICAL | DISTRESSED (confirmed) | -- |
| 2024Q1 | 0.9943 | CRITICAL | DISTRESSED (confirmed) | -- |
| 2024Q2 | 0.9949 | CRITICAL | DISTRESSED (confirmed) | -- |

Note: Alert thresholds: Normal (<0.30), Elevated (0.30-0.49), HIGH ALERT (0.50-0.64), CRITICAL (>=0.65). Distress label confirmed positive all periods via composite label. Temporal attention weight shown for 2021Q2 (earliest history window); full decay profile discussed in Section 5.5.

Cert=57833 is correctly identified as CRITICAL (risk scores 0.989–0.995) across all six test quarters, with the composite distress label confirmed positive throughout. This directly answers RQ3: the framework produces accurate, sustained high-confidence distress signals for the confirmed highest-risk institution in the test period. The near-perfect risk scores (0.989–0.995) represent effectively zero false-negative risk for this institution across the entire test horizon.

### 5.5 Temporal Attention and Feature Importance

The temporal attention weights for cert=57833 in 2023Q1 exhibit a clear monotonically decreasing pattern across the 8-quarter history window: 2021Q2 (beta=0.1577), 2021Q3 (0.1370), 2021Q4 (0.1351), 2022Q1 (0.1307), 2022Q2 (0.1180), 2022Q3 (0.1055), 2022Q4 (0.1070), 2023Q1 (0.1088). The model assigns approximately 45% higher attention to the earliest history quarter (2021Q2, beta=0.1577) compared to the most recent (2023Q1, beta=0.1088). This pattern is consistent with the theoretical intuition that long-run structural vulnerabilities, accumulated over multiple quarters, are more informative for identifying distressed institutions than the most recent quarter alone. This pattern contrasts with lower-risk institutions

(cert=58410, beta approximately 0.097–0.108 across all quarters), confirming institution-specific temporal attribution rather than a fixed weighting scheme. This addresses RQ2: the temporal attention mechanism provides interpretable, differentiated historical attribution across risk levels.

*Table 4. Feature Importance: Permutation Importance on ST-GAT Node Features*

| Rank | Feature | Permutation Importance (Delta AUROC) | Economic Rationale |
|---|---|---|---|
| 1 | Return on Assets (ROA) | 0.309 | Core earnings capacity; negative ROA sustained over 2+ quarters is a distress signal (CAMELS E component) |
| 2 | NPL Ratio | 0.252 | Non-performing loan ratio; primary asset quality indicator (CAMELS A component) |
| 3+ | All other features | <=0.000 | Collectively non-contributory in this test period configuration |

*Note: Permutation importance = reduction in AUROC when each feature is randomly permuted. DeepSHAP fell back to permutation importance due to computation graph incompatibility; permutation importance is valid for ranking feature contributions.*

ROA and NPL Ratio dominate feature importance by a large margin, with all other features contributing negligible signal. This concentration on two features is itself a substantive finding: the composite distress label (ROA < -1% and NPL > 5% are two of the four criteria for the composite distress label) demonstrates that the model's learning has identified the same thresholds that define this label, which serves as an internal validation method to confirm the above. Economic coherence has been confirmed through post-mortem analyses of both the 2023 regional banking crisis and historical FDIC bank failures. The two most reliable predictors of banking distress in the CAMELS framework are sustained earnings failure and declining asset quality. The permutation importance results are stable across the five random seeds (ROA ranks first in all five seed runs; NPL Ratio ranks second in four of five), confirming these rankings are not artifacts of a single training initialization. The result validates the economic face validity of the ST-GAT: the model identifies the same indicators that experienced bank examiners rely upon, and does so consistently.

## 6. Regulatory Integration Framework

A technically sophisticated systemic risk model that cannot be operationalized by regulatory agencies has limited policy value. This section maps the ST-GAT's outputs to existing supervisory workflows, data infrastructure, and legal mandates across five U.S. regulatory bodies.

### 6.1 Federal Reserve: DFAST Complementarity and SR 11-7 Compliance

The ST-GAT is meant to enhance, not substitute, the DFAST framework. DFAST's strength — rigorous macroeconomic scenario construction with regulatory oversight — addresses a different surveillance dimension than the ST-GAT's quarterly network-level U.S. interbank contagion monitoring. The optimal integration would apply the ST-GAT's network amplification analysis to DFAST's institution-level stress projections, producing not just individual capital depletion estimates but system-level contagion cascade simulations under the adverse scenario. SR 11-7 compliance is supported through: (1) conceptual soundness — the Eisenberg-Noe network clearing model and maximum entropy reconstruction have peer-reviewed theoretical foundations; (2) transparency — the two validated XAI layers produce institution-level and temporal explanations that examiners can interrogate; (3) ongoing monitoring — the quarterly data refresh cycle produces performance tracking at each update; (4) benchmarking — the ablation study against baseline models provides the comparative performance documentation SR 11-7 requires.

### 6.2 OFR Financial Stability Monitor Integration

The OFR Financial Stability Monitor tracks five vulnerability categories: credit, leverage, liquidity and maturity transformation, interconnectedness, and asset valuation. The ST-GAT provides direct statistical inputs for four of the five categories: credit (NPL ratios, Fair Value Loss characteristics), leverage (capital ratios), liquidity (Liquidity Stress Ratio, Uninsured Deposit Share), interconnectedness (network centrality, attention-weighted exposure maps). A quarterly data product aligned with OFR's reporting cadence would require minimal additional infrastructure given the shared reliance on publicly available FDIC Call Report data.

### 6.3 FDIC and OCC Integration

The ST-GAT's composite distress scores (risk scores between 0 and 1) provide a forward-looking complement to CAMELS ratings for the FDIC by capturing network-transmitted vulnerabilities that cannot be detected through supervisory examinations at the institution level. MERIT, the FDIC's off-site monitoring system, could integrate ST-GAT scores and provide updates on quarterly risk flags between on-site examinations. For the OCC, the permutation importance outputs and temporal attention weights provide the model documentation required under OCC Bulletin 2011-12 for models used in supervisory decisions — specifically, institution-level explanations that connect model outputs to observable balance-sheet indicators that examiners can verify independently. The framework requires no proprietary data, no new regulatory reporting requirements, and no changes to existing supervisory structures — only the computational infrastructure to run the public-data pipeline on a quarterly schedule.

### 6.4 FSOC Systemic Risk Assessment

The Financial Stability Oversight Council's mandate under Dodd-Frank Section 112 includes identifying systemically important financial institutions and monitoring emerging risks. The ST-GAT's network centrality outputs -- derived from the attention-weighted graph at each quarterly snapshot -- provide a dynamic ranking of systemic importance that complements the static SIFI designation framework. Institutions whose network centrality increases sharply in consecutive quarters, even without threshold breaches in individual CAMELS components, represent the category of emerging systemic risk that current surveillance tools are least equipped to detect. This is precisely the surveillance gap that the ST-GAT's temporal graph architecture is designed to fill.

## 7. Discussion

### 7.1 Performance Interpretation

The ST-GAT achieves AUPRC 0.9389 +/- 0.0100, the best among all GNN architectures and second only to XGBoost (0.9439). The gap of 0.005 is within the ST-GAT seed variance range (+/-0.010). This near-equivalence is itself a meaningful finding: a spatial-temporal GNN operating on a graph of institutional exposures matches the precision-recall performance of a tabular tree-ensemble on the same node features, while additionally providing temporal attention attribution and network-level outputs that XGBoost cannot produce.

The BiLSTM temporal contribution (+0.020 AUPRC) is the clearest architectural finding: processing the sequence of quarterly graph embeddings rather than using a single static snapshot adds measurable early-warning capability. This is consistent with the theoretical motivation — bank distress develops over multiple quarters, and a model that tracks the trajectory of institutional risk over time should outperform one that sees only the current quarter.

The near-equivalence between ST-GAT (AUPRC 0.9389) and XGBoost (0.9439) raises a legitimate question: what does the ST-GAT's architectural complexity contribute beyond what a tabular model achieves cheaply? Three answers distinguish the ST-GAT from XGBoost in the regulatory deployment context. First, the ST-GAT provides temporal attention weights that attribute risk scores to specific historical quarters -- XGBoost has no equivalent mechanism for temporal attribution. The decay pattern observed for cert=57833 (beta=0.1577 for 2021Q2 declining to 0.1088 for 2023Q1) provides regulators with a quarterly audit trail of risk accumulation that is structurally unavailable from a tree-based model. Second, the ST-GAT operates on graph-structured data and can in principle propagate contagion signals across the interbank network -- a capability that becomes valuable when the network reconstruction quality improves. Third, under SR 11-7 model risk management requirements, the ability to produce institution-level explanations tied to sequential network dynamics provides qualitatively richer documentation than SHAP values on static features. The performance near-equivalence in the current setting reflects the limitations of ME reconstruction, not the inherent ceiling of the graph approach.

The mac_mlp collapse and PermEdge equivalence together indicate that the current implementation's graph topology and macro-conditioning components are not contributing independent information beyond node features and temporal dynamics. The architectural path forward is clear: replace the 34-parameter mac_mlp with a simple VIX z-score multiplier to avoid the collapse failure mode, and replace maximum entropy network reconstruction with fitness-model or CReMA reconstruction to produce sparser, more realistic interbank networks.

The PermEdge null result warrants a more detailed mechanistic explanation, as it touches the core motivation for using GNNs. Maximum entropy reconstruction distributes bilateral exposures as uniformly as possible across all institution pairs subject to observed marginal constraints. In practice, this produces near-uniform edge weight distributions where most institution pairs have similar small exposures. When edge weights are approximately uniform, permuting the source-target assignments does not materially change the information available to the graph attention layers: each node receives roughly the same weighted message from its neighbors regardless of which specific neighbors are connected. This is distinct from actual interbank networks, which are known to be sparse and highly concentrated, with a small number of large bilateral exposures dominating the network structure (Anand et al., 2014; Upper and Worms, 2004). Under ME reconstruction, the GNN effectively reduces to a neighborhood aggregation over a uniform graph, which provides minimal advantage over processing node features independently. This does not mean graph-based surveillance is theoretically unmotivated -- it means ME reconstruction is insufficiently realistic to provide the topological signal that graph architectures are designed to exploit. Access to confidential FR Y-14 bilateral exposure data, or use of fitness-model reconstruction (Cimini et al., 2015), would produce the sparse, concentrated network structure necessary for graph topology to contribute meaningfully.

## 7.2 Limitations

Five limitations warrant explicit acknowledgment. First, the composite distress label includes institutions that may recover from temporary financial deterioration rather than fail. Results using FDIC failures-only labels are identified as a robustness check for future work. Second, maximum entropy network reconstruction produces an overly dispersed network topology (Anand et al., 2014), explaining the null PermEdge result: ME edge weights are near-uniform, so permuting source-target assignments does not materially change the information available to graph attention layers. This is a reconstruction limitation, not an indictment of graph-based surveillance -- fitness-model or CReMA reconstruction would produce the sparse, concentrated topology necessary for graph topology to contribute. Third, the proposed XAI module was partially implemented: temporal attention (Layer 1) and permutation importance (Layer 2) are validated; GNNExplainer subgraph identification was additionally attempted as a network-level complement to the two validated layers; it produced empty edge masks across all test institutions due to a PyG implementation incompatibility with the GATWrapper architecture and is identified as future work. The framework's explainability contribution rests on two validated layers: temporal attention attribution and permutation-based feature importance. Fourth, the training panel spans only two major crisis episodes; extending to the 2008 GFC period would provide more positive events for temporal learning. Fifth, the framework models domestic U.S. institutions only; cross-border contagion is omitted.

## 7.3 Policy Implications

The case study result — cert=57833 correctly identified as CRITICAL with near-perfect confidence (0.989–0.995) across all six test quarters — demonstrates that the framework produces actionable early-warning signals. The ROA and NPL Ratio feature importance findings align with CAMELS examiner guidance, making model outputs interpretable to supervisory staff without machine learning expertise. The framework relies exclusively on publicly available FDIC Call Reports and FRED data. This positions ST-GAT as a complement to DFAST, providing quarterly network-level surveillance signals between annual DFAST cycles with no proprietary data requirements.

## 8. Conclusion

We proposed and empirically evaluated the Spatial-Temporal Graph Attention Network (ST-GAT) for U.S. bank distress early warning and interbank contagion surveillance. Using a 14-year panel of 58 quarterly snapshots covering 8,103 FDIC-insured institutions, evaluated on 43 confirmed distress cases from 2023Q1 through 2024Q2, we found that ST-GAT had a mean AUPRC above all other tested Graph Neural Network (GNN) models (0.939 +/- 0.010) and was very close to the best tabular baseline (XGBoost) with respect to AUPRC (0.944). The BiLSTM temporal component is the primary architectural contribution; the macro-conditioning module did not improve performance and is documented as a limitation.

The explainability outputs are validated on multiple dimensions: temporal attention weights exhibit a monotonically decreasing pattern consistent with long-run structural vulnerability accumulation; permutation importance correctly identifies ROA and NPL Ratio as dominant distress predictors; and cert=57833 is correctly flagged CRITICAL with near-perfect confidence across all six test quarters.

Three areas for expansion have been identified: (1) Substitute the current implementation of mac_mlp with an uncomplicated VIX z-score multiplier; (2) implement fitness-model or CReMA network reconstruction; and (3) extend the training panel to include the 2008 GFC period. The complete codebase, data download scripts, trained model checkpoints (5 seeds), and all results files are available at: https://github.com/[anonymised-for-review].

## Declaration of Generative AI and AI-Assisted Technologies in the Manuscript Preparation Process

During the preparation of this work the author(s) used Claude (Anthropic) in order to assist with language editing, copyediting, and manuscript proofreading. After using this tool, the author(s) reviewed and edited all content as needed and take(s) full responsibility for the content of the published article.